\documentclass{article}
\usepackage[final]{colm2026}
\usepackage[colm,subcaption]{definition}

\usepackage{fontawesome5}
\usepackage{algorithm}
\usepackage[noend]{algpseudocode}
\usepackage{tcolorbox}
\tcbuselibrary{breakable}

\newcommand{\affilmark}[1]{\textsuperscript{\normalfont #1}}
\newcommand{\ours}{\textsc{RetroAgent}\xspace}
\newcommand{\hflogo}{\raisebox{-0.4ex}{\includegraphics[height=1em]{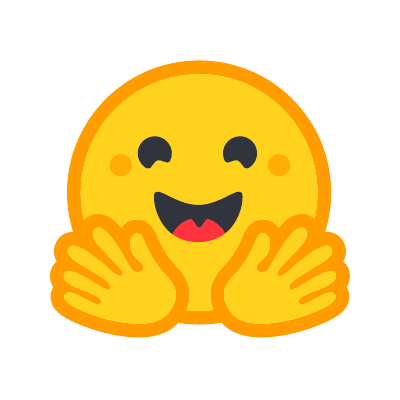}}}

\annotator{yanqiao}{red}
\annotator{jingru}{blue}

\title{\ours: Harnessing LLMs to Search Over Structured Memory for Agentic Retrosynthesis Planning}

\author{%
\normalsize
Yanqiao Zhu\affilmark{1,}$^{\ast}$\enspace
Jingru Gan\affilmark{1,}$^{\ast}$\enspace
Xiaoqi Sun\affilmark{2}\enspace
Fang Sun\affilmark{1}\enspace
Yidan Shi\affilmark{1}\enspace
Md Mofijul Islam\affilmark{3}\\
\normalsize
\textbf{%
Chao Shang\affilmark{3}\enspace
Wenhao Gao\affilmark{4}\enspace
Connor W. Coley\affilmark{2}\enspace
Yizhou Sun\affilmark{1}\enspace
Wei Wang\affilmark{1}}\\
\normalsize
\affilmark{1}UCLA \quad \affilmark{2}MIT \quad \affilmark{3}Amazon \quad \affilmark{4}UPenn \quad $^{\ast}$Equal contribution
\\
\normalsize\rule{0pt}{1em}
\faEnvelope[regular]{} Contact: \texttt{\{yzhu,jrgan\}@cs.ucla.edu}
}

\begin{document}

\maketitle

\begin{abstract}
Multi-step retrosynthesis planning seeks to decompose a target molecule into commercially available building blocks through a sequence of feasible reactions.
The vast combinatorial search space makes this task challenging even for expert chemists.
Traditional methods combine tree search with offline-trained value networks that score candidates in isolation, without reasoning about complete multi-step routes.
Recent work leverages Large Language Models (LLMs) for this task, but relies on simple interfaces that limit exploration of the full search space.
We introduce \ours, an LLM agent that bridges symbolic search and neural reasoning through a harness with structured memory.
Through memory and chemistry tools, the agent observes the full search state, including explored routes, available alternatives, and properties of intermediates, enabling informed decisions grounded in both global progress and domain knowledge.
Experiments on in-distribution and out-of-distribution benchmarks demonstrate that \ours delivers strong performance and generalization.

%\normalsize\rule{0pt}{1em}
\begin{center}
	\faGithub[]{}~Code: \url{https://github.com/SXKDZ/RetroAgent}
	
	\hflogo~Model: \url{https://huggingface.co/SXKDZ/RetroAgent}
\end{center}
\end{abstract}

\section{Introduction}
\label{sec:intro}

\begin{figure}
	\centering
	\vskip-0.5em
	\includegraphics[width=\linewidth]{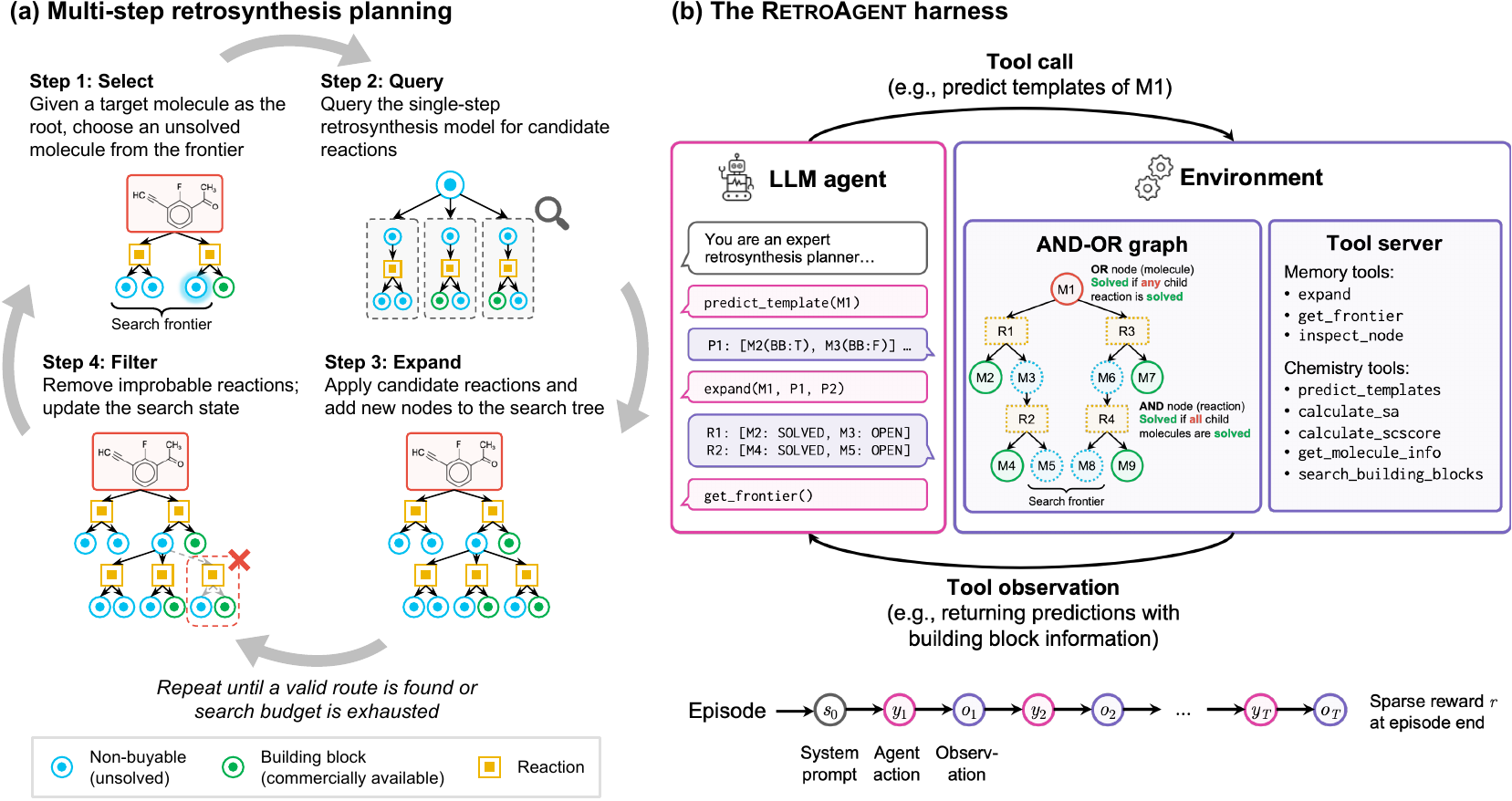}
	\caption{
	\textbf{(a)} Multi-step retrosynthesis planning iteratively decomposes a target molecule into commercially available building blocks through repeated cycles of selection, prediction, expansion, and filtering.
	\textbf{(b)} The harness of \ours, which couples an LLM agent with a structured-memory environment. The search proceeds over an AND-OR graph where molecule nodes are OR nodes, solved if any child reaction succeeds, and reaction nodes are AND nodes, solved only if all child molecules are solved. The agent calls memory tools to expand the tree and inspect the frontier, and chemistry tools to query the single-step model and obtain heuristic guidance. The environment maintains the graph, automatically propagates node statuses, and returns observations after each tool call. The agent is trained end-to-end with RL using an outcome-based reward.}
	\vskip-0.5em
	\label{fig:overview}
\end{figure}

Retrosynthetic analysis, the process of identifying synthetic routes to a target molecule from commercially available starting materials, is central to chemical research and drug development \cite{corey1969computer}.
Modern computational retrosynthesis decomposes this problem into two components \cite{zhong2024retrosynthesis,dong2022retrosynthesis, jiang2023retrosynthesis, tu2025askcos}: a single-step retrosynthesis model that predicts candidate reactions for a given molecule, and a multi-step planner that orchestrates the overall search by iteratively calling the predictor.
As illustrated in \cref{fig:overview}(a), given a target molecule, the planner iteratively selects a frontier molecule, queries the single-step model for candidate reactions, expands the search tree, and filters invalid candidates until a valid route is found.
The search space grows combinatorially with route depth, making multi-step retrosynthesis an inherently long-horizon planning task: decisions made early constrain downstream options, and a suboptimal choice can lead the search into regions where no feasible route exists.

For the single-step component, template-based models \cite{segler2017neural} classify over a library of reaction templates, while template-free approaches \cite{liu2017retrosynthetic,schwaller2019molecular,tetko2020state, tu2022graph2seq} generate product-to-reactant mappings directly.
For multi-step planning, research has advanced along two main themes: improving the single-step model through self-play on successful trajectories \cite{kim2021retro}, and training value functions to better guide the search process \cite{chen2020retro,liu2023pdvn,hong2023egmcts,segler2018planning,kishimoto2019dfpn}.
Reinforcement Learning (RL) has also been applied to learn expansion policies and value functions from simulated planning experience \cite{schreck2019learning,zhang2023rl}.
Despite significant progress, these methods rely on value functions or heuristics that are trained independently as proxy objectives rather than directly optimized for route completion.
Recently, Large Language Models (LLMs) have been applied as retrosynthetic planners.
\citet{wang2025llmsynplanner} use the LLM as a template-free route generator within an evolutionary pipeline augmented with retrieval.
Retro-R1 \cite{liu2025retror1} trains an LLM agent via RL to interactively construct routes, but its two-action interface provides limited flexibility to explore alternative decompositions.

In the broader literature, LLMs have shown increasing reasoning and planning capabilities, from linear reasoning chains \cite{wei2022chain} to environment-grounded action generation \cite{yao2023react} and tree search over reasoning states \cite{yao2023tree}.
LATS \cite{zhou2024language} takes a further step by embedding Monte Carlo tree search, which balances exploration and exploitation by building a search tree of possible actions, over LLM agent trajectories with environment feedback.
This demonstrates that LLMs benefit from operating within \emph{an explicit search structure with environmental grounding}, rather than relying solely on internal reasoning.
Our work brings this insight to retrosynthesis: rather than using the LLM as a value estimator within a fixed search algorithm, or as an unstructured sequential planner, we give the LLM direct control over a structured search process through tool calls.
Unlike traditional value functions that score molecules in isolation, an LLM agent can condition decisions on the full search context, enabling adaptive, state-dependent planning.

We introduce \ours, an LLM agent for retrosynthesis planning built on a harness that couples the agent with a structured-memory environment, as shown in \cref{fig:overview}(b).
The \textbf{environment} maintains an explicit search tree over an AND-OR graph, where OR-nodes represent molecules that can be decomposed in multiple ways and AND-nodes represent reactions that require all reactants to be solved.
It automatically propagates node statuses and computes the search frontier, while every reaction is grounded in a validated template set, ensuring that the search state is chemically valid.
The \textbf{agent} interacts with this environment through tool calls, choosing between memory tools that advance the search and chemistry tools that provide heuristic guidance.
This cleanly separates deterministic \textbf{state management}, such as status propagation and frontier computation, from \textbf{strategic decisions} that require reasoning, such as which molecule to expand and which reaction to apply. The environment thus guarantees a correct search state, while the agent focuses on learning effective search strategies.

Unlike previous approaches that either expand all candidates \cite{chen2020retro,liu2023pdvn} or commit to exactly one reaction \cite{liu2025retror1}, our harness lets the agent learn its own search strategy, including which molecule to prioritize, which and how many reactions to apply, all based on the current search context.
We train this agent end-to-end with GSPO \cite{zheng2025gspo} on retrosynthesis targets from the USPTO reaction database \cite{chen2020retro}, using a verifiable, outcome-based reward.
The reward further incorporates shaping terms that steer the learned behavior: for example, penalizing route depth encourages shorter routes, while rewarding expansion breadth encourages diverse decomposition pathways.

We evaluate \ours on the in-distribution USPTO-190 \cite{chen2020retro} and the out-of-distribution ChEMBL-1000 benchmarks \cite{liu2023pdvn}.
On USPTO-190, our agent achieves the highest pass@1 success rate among all methods with a smaller model than Retro-R1 and without relying on model probabilities to rerank predictions, while remaining competitive at larger budgets.
The advantage is most pronounced on ChEMBL-1000, where \ours surpasses all state-of-the-art methods on both pass@1 success rate and success rate at larger budgets.
This strong out-of-distribution performance suggests that the harness enables generalizable planning rather than dataset-specific memorization.

\section{Method}
\label{sec:method}

\subsection{Problem Formulation}
\label{sec:problem}

Given a target molecule $m_0$ and a set of commercially available building blocks $\mathcal{B}$, the goal of multi-step retrosynthesis planning is to find a synthesis route that decomposes $m_0$ into molecules in $\mathcal{B}$ through a sequence of valid reactions.
Each reaction applies a retrosynthesis template from a template library $\mathcal{T}$ to decompose a product molecule into precursors; a single-step retrosynthesis model $f$ proposes ranked candidate templates for any given molecule.
A valid route forms a directed acyclic graph from $m_0$ to building blocks, where every non-building-block molecule has at least one valid decomposition and the target is never consumed as a reactant.

The solution space of multi-step retrosynthesis can be naturally represented as an AND-OR graph $\mathcal{G} = (\mathcal{V}_M \cup \mathcal{V}_R, \mathcal{E})$: molecule nodes $\mathcal{V}_M$ have OR semantics, meaning a molecule is solved if any child reaction is solved, and reaction nodes $\mathcal{V}_R$ have AND semantics, meaning a reaction is solved only if all child molecules are solved.
Starting from the target molecule $m_0$ as the root, the search proceeds by iteratively expanding frontier molecules with candidate reactions from the single-step model, adding reaction nodes and edges $\mathcal{E}$ connecting molecules to their child reactions and reactions to their child molecules.

\subsection{Environment and Tool Interface}
\label{sec:environment}

We design an environment where the agent drives the search by calling tools that query and modify an AND-OR graph.
The environment maintains the graph as a structured memory that persists across the agent's interaction turns: each node carries a status $\sigma \in \{\texttt{Open}, \texttt{Solved}\}$, where a molecule starts as \texttt{Open} and becomes \texttt{Solved} once it is identified as a building block or successfully decomposed.
The set of \texttt{Open} leaf molecules forms the search frontier.

The agent interacts with this memory through two categories of tools.
Memory tools advance the search: the agent retrieves candidate reactions for a frontier molecule, selects which and how many to apply to grow the graph, and inspects the updated frontier to decide which molecule to expand next.
Chemistry tools provide auxiliary heuristics such as synthetic accessibility and building block availability for prioritizing among candidates.
We provide the full tool specifications in \cref{app:tools}.

The environment automatically handles status propagation, cycle detection, and frontier computation after each tool call, so the agent never needs to reason about graph consistency and can focus entirely on search strategy.
Through this interface, the agent learns to adapt its entire search strategy, including which molecule to prioritize, how many reactions to apply, and when to explore alternatives, rather than following the fixed policies of previous approaches \cite{chen2020retro,liu2025retror1}.

{\setlength{\textfloatsep}{8pt}%
\algrenewcommand\algorithmicindent{1em}%
\begin{algorithm}[t]
\caption{Expansion and Status Propagation}
\label{alg:propagation}
\small
\begin{minipage}[t]{0.47\textwidth}
\begin{algorithmic}[1]
\Require Graph $\mathcal{G}$, building blocks $\mathcal{B}$
\Function{Expand}{$\mathcal{G}$, molecule $m$, reactants $\{r_1, \ldots, r_k\}$}
    \State Create reaction node $R$
    \State Add $(m, R)$ to $\mathcal{E}$
    \For{each $r_i \in \{r_1, \ldots, r_k\}$}
        \State Look up $r_i$ in $\mathcal{V}_M$; create if absent
        \State Add $(R, n_i)$ to $\mathcal{E}$
        \If{$r_i \in \mathcal{B}$} $\sigma(n_i) \gets \texttt{Solved}$
        \EndIf
    \EndFor
    \State \Call{PropagateUp}{$R$}
\EndFunction
\end{algorithmic}
\end{minipage}%
\hfill
\begin{minipage}[t]{0.52\textwidth}
\begin{algorithmic}[1]
\Function{PropagateUp}{node $v$}
    \If{$v \in \mathcal{V}_R$} \Comment{AND semantics}
        \If{$\forall\, c \in \text{children}(v)\!: \sigma(c) {=} \texttt{Solved}$}
            \State $\sigma(v) \gets \texttt{Solved}$
        \EndIf
    \ElsIf{$v \in \mathcal{V}_M$} \Comment{OR semantics}
        \If{$\exists\, c \in \text{children}(v)\!: \sigma(c) {=} \texttt{Solved}$}
            \State $\sigma(v) \gets \texttt{Solved}$
        \EndIf
    \EndIf
    \If{$\sigma(v)$ changed}
        \For{each parent $u$ of $v$}
            \State \Call{PropagateUp}{$u$}
        \EndFor
    \EndIf
\EndFunction
\end{algorithmic}
\end{minipage}
\end{algorithm}}%

With this interface, the search becomes a sequential decision-making problem.
The graph is initialized with the target $m_0$ as the sole \texttt{Open} node.
At each turn, the agent can flexibly invoke tools to query candidate decompositions, expand one or more molecules, inspect node details, or call chemistry tools for heuristic guidance.
The multi-turn interaction between agent responses and environment observations forms a complete trajectory that we train end-to-end with RL.
The search terminates successfully when a complete route is found, or fails when the budget is exhausted.

\begin{wrapfigure}{r}{0.35\textwidth}
	\centering
	\vskip-0.2em
	\includegraphics[width=0.35\textwidth]{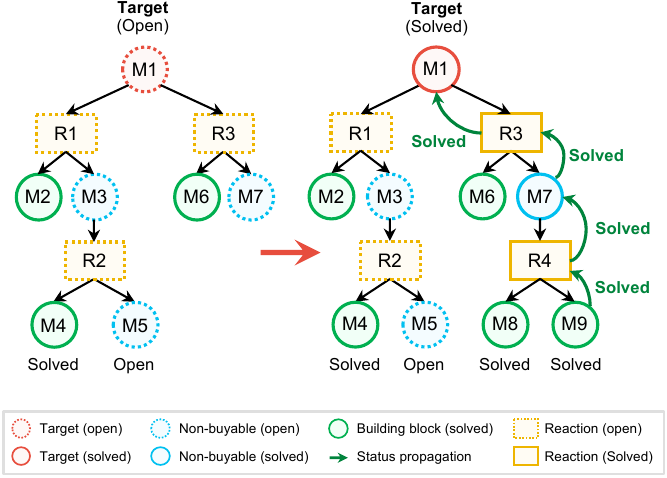}
	\caption{Example of status propagation: Expanding M7 with R4 triggers upward propagation through R4 $\to$ M7 $\to$ R3 $\to$ M1, solving the root node.}
	\label{fig:propagation}
\end{wrapfigure}
When the agent chooses to expand a molecule, the environment executes the expansion and status propagation (\cref{alg:propagation}): it adds new reactant nodes, checks each against $\mathcal{B}$, marks building block matches as \texttt{Solved}, and propagates status upward, where a reaction node becomes \texttt{Solved} when all children are solved following AND semantics, and a molecule node becomes \texttt{Solved} when any child reaction is solved following OR semantics.
Propagation only visits ancestors of newly changed nodes, requiring $O(d)$ time where $d$ is the route depth; route completion is then determined by a single $O(1)$ check of the root node's status.
\cref{fig:propagation} shows an example: Expanding M7 via R4 where both reactants M8 and M9 are building blocks triggers upward propagation through R4, M7, R3, and ultimately solves the root M1.

\textbf{Remarks.}
We highlight several design choices in the tool interface.
First, the agent can apply one or multiple candidate reactions in a single call, creating several OR branches at once.
Existing search-based methods expand with all available candidates \cite{chen2020retro,liu2023pdvn}, while Retro-R1 \cite{liu2025retror1} commits to exactly one reaction per molecule.
Our multi-expand design allows the agent to balance between these extremes: it can explore multiple promising reactions to hedge against dead ends while skipping clearly unpromising candidates, all within a single tool call.
Second, candidate decompositions returned by the single-step model include building block annotations for each reactant.
This allows the agent to immediately assess how close each reaction gets to completion without additional tool calls, reducing episode length and providing a direct signal for prioritizing candidates.
Third, the environment pre-filters single-step model predictions that would create cycles or duplicate existing reactions.
Since LLMs are unreliable at tracking graph structure \cite{wang2023nlgraph}, offloading validity checks to the environment ensures that the agent only sees valid, non-redundant candidates and can focus entirely on strategic evaluation.
Finally, the environment enforces early termination when the agent makes no progress for several consecutive turns or when all frontier molecules have no remaining viable predictions, preventing wasted computation on unsolvable targets.

\subsection{Training with Reinforcement Learning (RL)}
\label{sec:training}

We use the LLM as the policy $\pi_\theta$ and train it end-to-end with RL using verifiable, outcome-based rewards.
Each episode is a complete multi-turn search trajectory; only agent tokens are trainable, while tool observations are masked.

\subsubsection{Reward Function}

The reward is verifiable and outcome-based: the agent receives a positive reward only when it finds a complete route to building blocks, with no partial credit for incomplete routes.
Beyond this binary outcome, we add reward shaping terms that encourage efficient search behavior by penalizing excessive budget usage and deep routes.
The reward at episode end is defined as:
\begin{equation}
\label{eq:reward}
    r = \begin{cases}
        1.0 - \delta_N - \delta_D + p, & \text{if route completed,} \\
        -1.5, & \text{if no valid tool call is ever made,} \\
        -1.0, & \text{otherwise,}
    \end{cases}
\end{equation}
where $\delta_N$ penalizes excessive search budget measured as the number of single-step model calls and $\delta_D$ penalizes deep routes.
The $-1.5$ case applies to degenerate trajectories that never emit a well-formed tool call (pure format failure), penalizing them more heavily than genuine but unsuccessful search attempts.

Both penalties follow the same form: zero below a threshold of 20, then 0.02 per unit above, capped at 0.3:
\begin{equation}
    \delta_N = \min\left(0.3,\; 0.02 \cdot \max(0,\, N - 20)\right), \quad
    \delta_D = \min\left(0.3,\; 0.02 \cdot \max(0,\, D - 20)\right),
\end{equation}
where $N$ is the number of single-step model calls and $D$ is the maximum route depth in reaction steps from the root.
The term $p$ accumulates minor penalties of $-0.05$ for turns with tool errors or no valid tool calls, discouraging the agent from generating malformed actions.
We note that this reward design is flexible: adjusting the relative weights of $\delta_N$ and $\delta_D$ induces different search behaviors, e.g., emphasizing $\delta_D$ encourages shorter routes while emphasizing $\delta_N$ encourages budget-efficient exploration.

\subsubsection{Policy Optimization}

We use GSPO \cite{zheng2025gspo}, a critic-free policy gradient method that extends GRPO \cite{shao2024deepseekmath} by applying clipping at the sequence level rather than per token.
The training objective combines a clipped surrogate policy loss $\mathcal{L}_{\text{GSPO}}$ with a KL regularization term $\mathcal{L}_{\text{KL}}$ against the frozen pretrained model.

For each batch, we sample $B$ target molecules and generate $G$ rollout episodes per target, indexed by $i$ and $g$ respectively.
Advantages are computed by normalizing rewards within each group, eliminating the need for a separately trained critic:
\begin{equation}
\label{eq:advantage}
    \hat{A}_{i,g} = \frac{r_{i,g} - \bar{r}_i}{\max(\text{std}(\bm{r}_i),\, \epsilon)},
    \qquad \text{where } \bar{r}_i = \frac{1}{G}\sum_{g=1}^G r_{i,g}.
\end{equation}
The policy loss uses a sequence-level clipped surrogate objective \cite{schulman2017proximal}:
\begin{equation}
\label{eq:gspo}
    \mathcal{L}_{\text{GSPO}}(\theta) = -\mathbb{E}_{i,g}\left[\min\left(s_{i,g}\,\hat{A}_{i,g},\;\text{clip}(s_{i,g}, 1{-}\epsilon, 1{+}\epsilon)\,\hat{A}_{i,g}\right)\right],
\end{equation}
where $s_{i,g} = \exp\!\left(\frac{1}{|\bm{y}_{i,g}|}\sum_t \log \frac{\pi_\theta(y_t \mid \bm{y}_{<t})}{\pi_{\text{old}}(y_t \mid \bm{y}_{<t})}\right)$ is the length-normalized sequence-level importance ratio, $t$ indexes tokens in the trajectory, and $\epsilon$ is the clip ratio.
To improve sample efficiency, we reuse each batch of rollouts for multiple gradient steps with Truncated Importance Sampling (TIS), clipping the sequence-level importance ratio to a fixed range and discarding samples outside this range to control off-policy variance.

The KL term regularizes against the frozen pretrained model $\pi_{\text{ref}}$ using the $k_3$ low-variance estimator \cite{schulman2020kl} to prevent reward saturation:
\begin{equation}
\label{eq:kl}
    \mathcal{L}_{\text{KL}}(\theta) = \beta \cdot \mathbb{E}_t\left[\frac{r_t - 1}{r_t} - \log r_t\right], \qquad r_t = \frac{\pi_\theta(y_t \mid \bm{y}_{<t})}{\pi_{\text{ref}}(y_t \mid \bm{y}_{<t})},
\end{equation}
where $\beta$ is the KL penalty coefficient.

\section{Experiments}
\label{sec:experiments}

We conduct experiments to evaluate \ours through the following questions:
\begin{itemize}
    \item \textbf{RQ1.} Does \ours outperform existing search-based and LLM-based methods on both in-distribution and out-of-distribution benchmarks?
    \item \textbf{RQ2.} Is RL training essential, or does the base model already perform well with the structured memory interface?
    \item \textbf{RQ3.} How do key design choices, specifically multi-expand and depth penalty, affect the agent's performance?
    \item \textbf{RQ4.} What search behavior does the RL-trained agent learn, and how does it differ from ablated variants?
\end{itemize}

\subsection{Experimental Setup}
\label{sec:setup}

\textbf{Datasets.}
Following prior work \cite{chen2020retro}, we use a building block set of 23,081,633 commercially available molecules from eMolecules.\footnote{\url{https://downloads.emolecules.com/}}
For RL training, we extract 20,626 target molecules from the USPTO-Full reaction database \cite{chen2020retro}, filtered to routes with depth greater than 4, with mean depth 5.93.
We evaluate on two benchmarks:
(1) USPTO-190: 190 challenging target molecules extracted from USPTO-Full \cite{chen2020retro} with reference routes (mean depth 6.0), serving as the standard in-distribution benchmark;
(2) ChEMBL-1000: 1,000 drug-like molecules from the ChEMBL database \cite{liu2023pdvn}, an out-of-distribution dataset without reference routes, used to test generalizability.

\textbf{Single-step model.}
We follow prior work \cite{chen2020retro} and use a template-based MLP as the single-step retrosynthesis model.
The model predicts over 381,302 USPTO reaction templates based on Morgan fingerprints, and uses RDChiral \cite{coley2019rdchiral} for template application and verification.
We use the pretrained weights from DESP \cite{yu2024desp}.
We do not adopt the Retro-R1 weight variants V2, V3, and V4 \cite{liu2025retror1}: V2 and V3 may produce chemically unrealistic reactions \cite{genheden2022paroutes}, and V4 weights are not public.
We also do not adopt their two-branch policy that reranks reactions based on predicted probabilities, as we aim to test the learned search capabilities of the agent.
Instead, we shuffle all prediction and frontier orderings during both training and evaluation, so the agent evaluates candidates by content rather than positional bias.

\textbf{Baselines.}
We compare with both search- and LLM-based baselines.
Greedy DFS always selects the reaction with the highest likelihood \cite{chen2020retro}.
Retro* \cite{chen2020retro} is an A*-like best-first search where the past cost is derived from single-step model probabilities and the future cost is estimated by a learned value network.
Retro*-0 \cite{chen2020retro} is its non-learning variant where the future cost is set to 0.
Retro*+ and Retro*+-0 \cite{kim2021retro} improve Retro* and Retro*-0, respectively, with self-improved single-step models.
EG-MCTS \cite{hong2023egmcts} improves Retro*+ with experience-guided Monte Carlo tree search.
PDVN \cite{liu2023pdvn} trains a single-step model with an online algorithm and uses dual value networks for synthesizability and cost estimation.
Furthermore, we compare against the state-of-the-art Retro-R1 \cite{liu2025retror1}, a \verb|Qwen2.5-7B-Instruct-1M| agent with a two-action interface that follows depth-first search without backtracking.

\textbf{Metrics.}
Following \citet{liu2025retror1}, we report two complementary metrics.
Pass@1 measures the fraction of targets solved in a single trajectory, directly reflecting the agent's decision quality without retry.
Success rate at budget $N$ measures the fraction of targets solved within $N$ unique single-step model calls, capturing cumulative performance as the search budget increases.
While the budget definition is the same for all methods, search-based methods consume the full budget within a single deterministic run.
LLM-based methods typically cannot exhaust the budget in one trajectory due to the difficulty of long-context modeling; we therefore follow the iterative evaluation protocol \cite{liu2025retror1} and conduct multiple independent trajectories per target with shuffled prediction and frontier orderings, where repeated predictions do not count toward the budget.
Details of our evaluation protocol are provided in \cref{app:eval-protocol}.

\textbf{Model and infrastructure.}
We use \verb|Qwen3-4B-Instruct-2507| \cite{yang2025qwen3} as the base model, a 4B-parameter instruction-tuned model with native tool-calling support.
We deliberately choose a smaller model without thinking capabilities to demonstrate that the structured memory interface, rather than raw model capacity, drives planning
performance.
Training uses the slime framework \cite{zhu2025slime} with Megatron \cite{shoeybi2019megatron} for distributed training and SGLang \cite{zheng2024sglang} for inference on
 $8{\times}$H200 GPUs.
The tool server runs as a separate process pool with 16 instances, decoupled from the training pipeline; this allows the environment to scale independently of GPU resourc
es and avoids blocking gradient computation during tool execution.

\textbf{Hyperparameter configuration.}
We optimize with Adam ($\beta_1=0.9$, $\beta_2=0.98$), a constant learning rate of $1{\times}10^{-6}$, and weight decay of 0.01.
Each batch samples 32 target molecules with $G=8$ rollouts each at temperature 1.0, yielding 256 episodes per batch; with TIS reusing each batch for 2 gradient steps, the effective batch size is 128.
The GSPO clip ratio is $\epsilon = 3.5{\times}10^{-4}$, the KL coefficient is $\beta = 0.001$, and the TIS clip range is $[0.5, 2.0]$.
Each training episode runs for up to 90 turns with a 56K token budget and a maximum route depth of 30 reaction steps.
At evaluation time, we use greedy decoding with temperature 0 to ensure deterministic runs, with more generous limits of 120 turns and 120K tokens.
For each molecule, we take the 50 highest-scoring predictions from the single-step model, shuffle them, and present at most 20 to the agent, removing positional bias so the agent judges candidates by content rather than rank.
Full hyperparameters and training curves are reported in \cref{app:training}.

\subsection{Main Results (RQ1)}
\label{sec:main-results}

\cref{tab:main} summarizes results on the in-distribution USPTO-190 and the out-of-distribution ChEMBL-1000.
Since Retro-R1 does not report its V1 results on ChEMBL-1000, we reproduce Retro-R1 using its released code and evaluate on this dataset.
All other performance results are directly from Retro-R1 \cite{liu2025retror1}.

\begin{table*}[t]
\centering
\caption{Pass@1 and success rate (\%) at different budgets $N$ on USPTO-190 and ChEMBL-1000. Best in \textbf{bold}, second best \underline{underlined}.}
\label{tab:main}
\begin{tblr}{
    colspec = {l l c cccccc},
	row{1-2} = {bg=gray!25, font=\bfseries},
	cell{4,6,8,10,12,14,16,18,20}{2-9} = {bg=gray!10}
}
	\toprule
    & \SetCell[r=2]{l} Method & \SetCell[r=2]{c}{Pass@1 \\ (\%)} & \SetCell[c=6]{c} Success Rate (\%) at Budget $N$ \\
    \cmidrule[lr]{4-9}
    & &  & $N=50$ & $N=100$ & $N=200$ & $N=300$ & $N=400$ & $N=500$ \\
    \midrule
    \SetCell[r=9]{l} \rotatebox{90}{USPTO-190}
    & Greedy DFS & 19.47 & 19.47 & 19.47 & 19.47 & 19.47 & 19.47 & 19.47 \\
    & Retro* & 20.53 & 38.95 & 50.00 & 68.95 & 74.74 & 77.89 & 79.47 \\
    & Retro*-0 & 19.47 & 27.37 & 37.89 & 55.79 & 62.63 & 72.11 & 74.21 \\
    & Retro*+ & 30.00 & 55.79 & 68.42 & 79.47 & \underline{83.16} & 84.21 & 84.21 \\
    & Retro*+-0 & 25.26 & 47.89 & 61.05 & 76.84 & 82.11 & \underline{85.79} & \underline{86.32} \\
    & EG-MCTS & 46.84 & 58.42 & 64.74 & 69.47 & 75.26 & 78.42 & 81.05 \\
    & PDVN & 42.63 & \textbf{64.74} & \textbf{72.11} & \textbf{82.63} & \textbf{87.89} & \textbf{91.58} & \textbf{92.11} \\
    & Retro-R1 (V1) & \underline{50.00}{\scriptsize$\pm$0.00} & \underline{64.63}{\scriptsize$\pm$1.68} & 71.32{\scriptsize$\pm$1.23} & \underline{78.11}{\scriptsize$\pm$1.68} & 80.89{\scriptsize$\pm$1.81} & 82.95{\scriptsize$\pm$1.40} & 84.32{\scriptsize$\pm$1.22} \\
    \cline[dashed]{2-9}
    & \ours & \textbf{53.26}{\scriptsize$\pm$0.69} & 58.42{\scriptsize$\pm$1.34} & 70.53{\scriptsize$\pm$1.93} & \underline{78.63}{\scriptsize$\pm$1.27} & 81.58{\scriptsize$\pm$0.83} & 83.68{\scriptsize$\pm$0.37} & 85.05{\scriptsize$\pm$0.80} \\
    \midrule
    \SetCell[r=9]{l} \rotatebox{90}{ChEMBL-1000}
    & Greedy DFS & 38.10 & 38.10 & 38.10 & 38.10 & 38.10 & 38.10 & 38.10 \\
    & Retro* & 47.70 & 65.60 & 69.00 & 71.60 & 73.20 & 74.00 & 74.40 \\
    & Retro*-0 & 38.10 & 64.60 & 67.20 & 70.30 & 71.80 & 72.70 & 73.40 \\
    & Retro*+ & 53.50 & 70.50 & 73.90 & 76.20 & 78.00 & 78.80 & 79.40 \\
    & Retro*+-0 & 49.00 & 69.40 & 73.60 & 75.60 & 76.70 & 77.60 & 78.50 \\
    & EG-MCTS & 59.70 & 71.20 & 74.20 & 76.90 & 78.30 & 79.00 & 79.30 \\
    & PDVN & 60.00 & 73.90 & 76.40 & 78.10 & 79.60 & 80.30 & 80.60 \\
    & Retro-R1 (V1) & \underline{68.52}{\scriptsize$\pm$1.58} & \underline{73.92}{\scriptsize$\pm$0.37} & \underline{78.00}{\scriptsize$\pm$0.46} & \underline{80.36}{\scriptsize$\pm$0.35} & \underline{81.36}{\scriptsize$\pm$0.27} & \underline{81.98}{\scriptsize$\pm$0.28} & \underline{82.42}{\scriptsize$\pm$0.22} \\
    \cline[dashed]{2-9}
    & \ours & \textbf{73.82}{\scriptsize$\pm$0.49} & \textbf{76.36}{\scriptsize$\pm$0.35} & \textbf{79.62}{\scriptsize$\pm$0.15} & \textbf{81.96}{\scriptsize$\pm$0.30} & \textbf{83.22}{\scriptsize$\pm$0.26} & \textbf{83.94}{\scriptsize$\pm$0.30} & \textbf{84.40}{\scriptsize$\pm$0.38} \\
    \bottomrule
\end{tblr}
\end{table*}

On USPTO-190, \ours achieves 53.26\% pass@1 and 85.05\% at $N = 500$ with a 4B-parameter model, outperforming Retro-R1 with 50.00\% pass@1 and 84.32\% at $N = 500$, despite using a $1.75\times$ smaller model of 4B versus 7B parameters and not relying on model probabilities to rerank predictions.
Search-based methods such as PDVN train their value networks on all $\sim$300K available routes in USPTO-Full, while \ours uses only 20K targets, highlighting its data efficiency.
At higher budgets ($N \geq 400$), the exhaustive search of PDVN surpasses LLM-based methods, reaching 92.11\% at $N = 500$.
It is important to note that this comparison is not on equal footing: search-based methods apply all available predictions per expansion, up to 50, whereas at evaluation on USPTO-190 our agent selectively applies a mean of 1.71 reactions per \verb|expand| call.
At the same budget $N$, search-based methods thus explore a substantially larger portion of the search space, which gives them an inherent advantage at high budgets.
Despite this, we achieve the highest pass@1 among all methods, demonstrating stronger single-trajectory decision quality.

Our advantage is most evident on the out-of-distribution ChEMBL-1000, where \ours achieves 73.82\% pass@1 and 84.40\% at $N = 500$, outperforming Retro-R1 by 5.30\% in pass@1 and 1.98\% at $N = 500$, and surpassing all search-based methods including PDVN (80.60\% at $N = 500$).
This suggests the structured memory interface helps the agent acquire transferable planning strategies rather than memorizing dataset-specific reaction patterns.
A detailed out-of-distribution breakdown is provided in \cref{app:ood}.

\subsection{Ablation Studies (RQ2, RQ3)}
\label{sec:ablation}

We ablate three aspects of our agent on USPTO-190: the necessity of RL training, the effect of multi-expand versus single-expand, and the impact of the depth penalty $\delta_D$.
All variants use the same evaluation protocol and tool interface; the zero-shot baseline uses temperature 0.6 since greedy decoding with the untrained model produces degenerate outputs.

\begin{table}[t]
\centering
\caption{Ablation studies on USPTO-190.}
\label{tab:ablation}
\small
\begin{tblr}{
    colspec = {l c cccc c},
	row{1-2} = {bg=gray!25, font=\bfseries},
	row{4,6} = {bg=gray!10}
}
    \toprule
    \SetCell[r=2]{l} Configuration & \SetCell[r=2]{c}{Pass@1 \\ (\%)} & \SetCell[c=4]{c} Success Rate (\%) at Budget $N$ & & & & \SetCell[r=2]{c}{Shorter \\ Routes} \\
    \cmidrule[lr]{3-6}
    & & $N=50$ & $N=100$ & $N=200$ & $N=500$ & \\
    \midrule
    \ours & \textbf{53.26}{\scriptsize$\pm$0.69} & \textbf{58.42}{\scriptsize$\pm$1.34} & \textbf{70.53}{\scriptsize$\pm$1.93} & \textbf{78.63}{\scriptsize$\pm$1.27} & \textbf{85.05}{\scriptsize$\pm$0.80} & \textbf{38.3}{\scriptsize$\pm$3.1} \\
    \midrule[dashed]
    Zero-shot (no RL) & 4.42{\scriptsize$\pm$1.27} & 15.89{\scriptsize$\pm$1.88} & 16.53{\scriptsize$\pm$1.88} & 16.63{\scriptsize$\pm$2.02} & 16.63{\scriptsize$\pm$2.02} & 16.4{\scriptsize$\pm$3.6} \\
    \midrule[dashed]
    w/o depth penalty & 51.16{\scriptsize$\pm$2.59} & 54.95{\scriptsize$\pm$1.42} & 64.84{\scriptsize$\pm$2.39} & 73.79{\scriptsize$\pm$1.01} & 81.26{\scriptsize$\pm$0.47} & 31.4{\scriptsize$\pm$3.6} \\
    w/o multi-expand & 40.63{\scriptsize$\pm$2.05} & 52.21{\scriptsize$\pm$1.95} & 63.37{\scriptsize$\pm$1.09} & 71.58{\scriptsize$\pm$1.49} & 79.05{\scriptsize$\pm$0.69} & 32.8{\scriptsize$\pm$2.6} \\
    w/o both & 39.58{\scriptsize$\pm$2.02} & 52.32{\scriptsize$\pm$1.52} & 62.84{\scriptsize$\pm$0.96} & 72.00{\scriptsize$\pm$1.36} & 77.37{\scriptsize$\pm$1.34} & 26.6{\scriptsize$\pm$1.7} \\
    \bottomrule
\end{tblr}
\end{table}

The results are reported in \cref{tab:ablation}. The zero-shot model achieves only 4.42\% pass@1, confirming that RL is essential and ruling out data leakage from pretraining.
Multi-expand is the most impactful design choice, improving pass@1 by over 12 points: from 40.63\% to 53.26\% with depth penalty, and from 39.58\% to 51.16\% without.
The depth penalty further raises success rate at $N = 500$ from 81.26\% to 85.05\%.
We also report ``Shorter Routes'' in \cref{tab:ablation}, the number of targets (out of 190) for which the found route is strictly shorter than the reference route, averaged over runs (mean$\pm$std). \ours finds the most shorter routes ($38.3\pm3.1$ per run), ahead of all ablated variants ($26.6$--$32.8$); removing the depth penalty or multi-expand reduces shorter-route discovery, and removing both reduces it most. Aggregating over runs, \ours produces a shorter route than the reference for 69 distinct targets in total.

To further isolate our method from the choice of base model, we reproduce the Retro-R1 two-action interface on our own \texttt{Qwen3-4B-Instruct-2507} backbone, holding the model and RL training fixed and varying only the interface. It reaches $45.05\pm0.29$\% pass@1 on USPTO-190 (five-run mean), $8.21$ points below \ours and even below the original Qwen2.5-7B Retro-R1 result of 50.00\%, confirming that the gain comes from the interface rather than the newer backbone. We also examine the sensitivity of our reward shaping: sweeping the relative weighting and magnitude of the budget and depth penalties moves pass@1 by only a few points, and a harsher penalty magnitude trades first-attempt success for a converging budget curve rather than removing the ability to solve targets, indicating that performance is not overly sensitive to these values. Please refer to \cref{app:additional} for full results of both studies.

\subsection{Behavioral Analysis (RQ4)}
\label{sec:behavior}

\begin{figure*}[t]
\centering
\includegraphics[width=\textwidth]{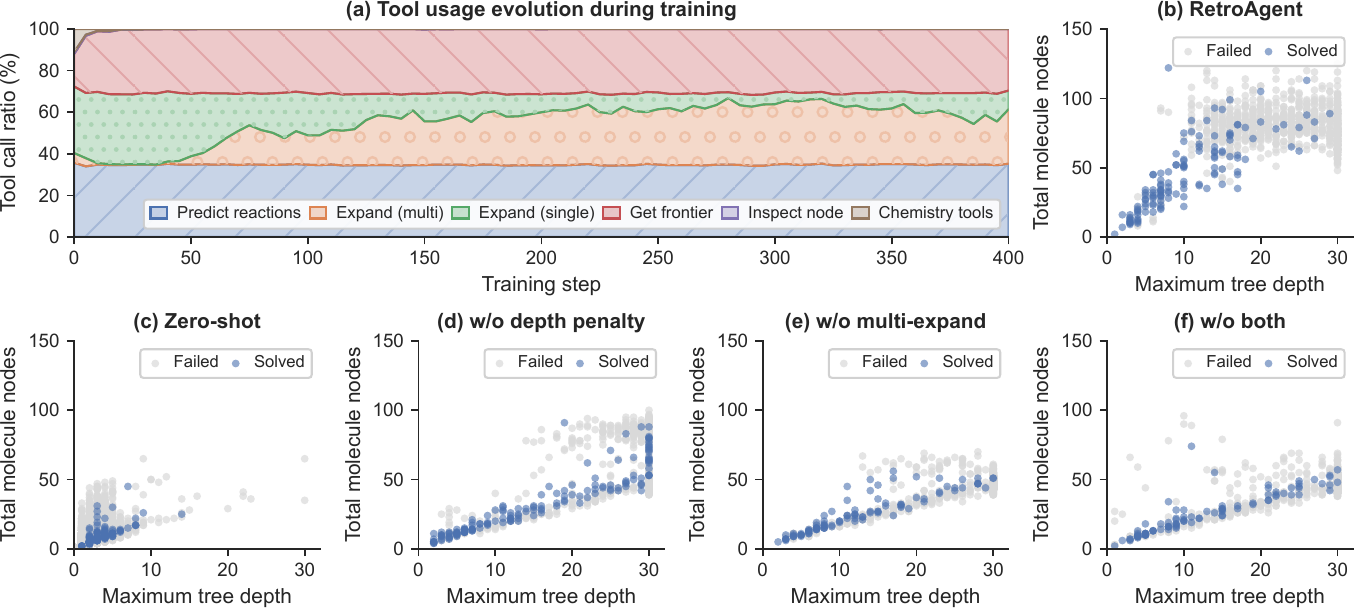}
\caption{Behavioral analysis of \ours and ablated variants on USPTO-190.
\textbf{(a)} Tool call ratio during RL training.
\textbf{(b)--(f)} Search tree structure for each configuration. Each point is one rollout; the $x$-axis is the maximum depth of the search tree in reaction steps, and the $y$-axis is the total number of molecule nodes expanded during the episode.}
\label{fig:ablation}
\end{figure*}

Beyond the ablation numbers, we analyze what search strategy the agent actually learns by examining tool usage evolution and search tree structure (\cref{fig:ablation}), which helps further explain the performance gaps in \cref{tab:ablation}.

\textbf{Tool usage evolution.}
\cref{fig:ablation}(a) shows how tool call ratios change during training.
The agent has access to 8 tools: 4 memory tools for search and 4 chemistry tools for heuristic guidance.
Chemistry tools are actively used at step 0 ($\sim$11\%) but collapse to near zero by step 10, suggesting that the building block annotations returned by the prediction tool provide sufficient signal for candidate prioritization.
The node inspection tool stays below $\sim$1\% throughout training and quickly drops to near zero, indicating the agent extracts sufficient information from other tool responses.
Three tools remain dominant throughout: prediction, expansion, and frontier retrieval.
Multi-expand usage emerges around step 120 and becomes dominant after step 350, with the agent applying multiple reactions per call in over 70\% of expansions, demonstrating that the agent discovers this strategy through reward shaping rather than being explicitly guided.

\textbf{Search tree structure.}
\cref{fig:ablation}(b)--(f) plot the search tree structure for each configuration, where each point represents one rollout with its maximum tree depth and total molecule nodes expanded.
Comparing across configurations reveals three key findings.

First, RL training is essential for long-horizon planning.
As shown in \cref{fig:ablation}(c), the zero-shot model clusters all rollouts in the lower-left corner, rarely exceeding 15 steps or 75 nodes regardless of outcome, indicating it cannot sustain productive search beyond the first few expansions.
This corroborates the success rate plateau beyond $N=100$ in \cref{tab:ablation}.
In contrast, the RL-trained agent in \cref{fig:ablation}(b) routinely builds trees with 50--100+ nodes, with solved cases concentrated at shallow depth.

Second, multi-expand trades depth for breadth.
As seen in \cref{fig:ablation}(e), removing multi-expand yields narrow trees with fewer nodes but greater depth to find solutions.
The full model in \cref{fig:ablation}(b) creates more nodes at shallower depth by expanding multiple OR branches per molecule, hedging against dead ends without committing to deep search.
This breadth-first tendency directly explains why multi-expand yields 12+ points of pass@1 improvement while also finding more shorter routes than any ablated variant.
The agent applies this breadth selectively rather than exhaustively: across all \verb|expand| calls, $\sim$84\% apply only one or two reactions ($w=1$: 54.3\% on USPTO-190 and 55.6\% on ChEMBL-1000; $w=2$: 29.3\% and 28.2\%), with a thin tail beyond $w=5$ ($\sim$0.6\% on both).
Solved targets expand slightly wider on average than failed ones (mean width 1.81 vs.\ 1.65 on USPTO-190; 1.86 vs.\ 1.66 on ChEMBL-1000), indicating that multi-expand contributes through selective local breadth at promising decision points rather than indiscriminate branching.

Third, depth penalty prevents inefficient deep exploration.
As illustrated in \cref{fig:ablation}(d), without the depth penalty, solved cases frequently extend to the maximum allowed depth of 30, wasting budget on unnecessarily long routes.
Removing both design choices as in \cref{fig:ablation}(f) yields the weakest configuration, though the agent can still solve some targets through single-expand depth-first search.

\section{Conclusion}
\label{sec:conclusion}

We presented \ours, an RL-trained LLM agent for retrosynthesis planning that operates over an AND-OR graph as structured memory.
The environment automatically maintains the search state, while the agent makes strategic decisions through tool calls.
With a 4B-parameter model, \ours achieves the highest pass@1 on USPTO-190 and substantially outperforms all baselines on the out-of-distribution ChEMBL-1000, despite using a smaller model and less training data.
Ablation studies confirm that multi-expand and reward shaping are key to the agent's performance, and behavioral analysis reveals that the agent progressively learns to trade depth for breadth through RL.
The strong out-of-distribution results suggest that the harness enables generalizable planning strategies rather than dataset-specific memorization.
We believe the design principle of combining LLM reasoning with structured environment management is broadly applicable to other long-horizon planning tasks beyond retrosynthesis.

\section*{Acknowledgments}
This work was partially supported by NSF (2106859, 2202693, 2303037, 2312501, 2531008, and 2609582), the SRC JUMP 2.0 Center, the UCLA CDSC Center, and the National Artificial Intelligence Research Resource (NAIRR) Pilot (240280, 240443).
We also gratefully acknowledge support from Amazon Research Awards, NEC, Snapchat, and Google Gifts.
Y.Z. acknowledges additional support from the Tinker Grant and the Lambda Research Grant during the early stages of this work.

\section*{LLM Usage Disclosure}
We disclose that LLMs were used for editing and polishing the language of this manuscript. All scientific content, experimental design, and analysis were conducted by the authors.

\bibliographystyle{plainnat}
\bibliography{references}

\newpage
\appendix
\appendixprefix
\vspace*{1pt}
\begin{center}
	{\Large \textbf{Supplementary Material for \ours}}
\end{center}
\vspace*{1pt}

\startcontents[appendix]
\printcontents[appendix]{}{1}{\setcounter{tocdepth}{2}}
\vspace{1em}

\section{System Prompt}
\label{app:prompt}

The system prompt instructs the agent on the AND-OR graph structure, the available tools, and the search loop.
The user prompt provides the target molecule SMILES and instructs the agent to begin with \verb|predict_templates("M1")|.

\begin{tcolorbox}[colback=gray!5, colframe=gray!50, title=System Prompt, fonttitle=\bfseries\small, fontupper=\small]
You are an expert retrosynthesis planner. Decompose a target molecule into commercially available building blocks through valid reactions.

\medskip
You search over an AND-OR graph. Molecules (M1, M2, \ldots) are OR nodes: a molecule can be decomposed in multiple ways, and any one successful decomposition suffices. Reactions (R1, R2, \ldots) are AND nodes: all reactants must be solved for the reaction to succeed. Status propagates automatically up the tree.

\medskip
Core loop:
\begin{enumerate}[nosep,leftmargin=*]
    \item \verb|predict_templates(molecule_id)|: get candidate reactions (P1, P2, \ldots)
    \item \verb|expand(molecule_id, prediction_id)|: apply predictions to the graph. Use comma-separated IDs (e.g., ``P1,P3'') to expand multiple at once.
    \item \verb|get_frontier()|: see which OPEN molecules still need decomposition.
    \item Repeat until the route is solved (automatic detection).
\end{enumerate}

\medskip
Maximum route depth: 30 reaction steps.
\end{tcolorbox}

\begin{tcolorbox}[colback=blue!3, colframe=blue!40, title=User Prompt (per target), fonttitle=\bfseries\small, fontupper=\small]
Plan a retrosynthesis route for: \verb|{target_smiles}|

\medskip
The target is M1. The graph tools (\verb|predict_templates|, \verb|expand|, \verb|inspect_node|) use molecule IDs (M1, M2, \ldots). The chemistry tools (\verb|get_molecule_info|, \verb|search_building_blocks|, \verb|calculate_synthetic_accessibility|, \verb|calculate_scscore|) use SMILES strings. Start with \verb|predict_templates("M1")|.
\end{tcolorbox}

\section{Tool Interface Details}
\label{app:tools}

The agent interacts with the environment through 8 tools divided into two categories.
All graph-related tools use symbolic identifiers (\verb|M#| for molecules, \verb|R#| for reactions, \verb|P#| for predictions) rather than raw SMILES strings to prevent copying errors.

\textbf{Memory tools.}
These tools allow the agent to query and modify the AND-OR graph:
\begin{itemize}[nosep,leftmargin=*]
    \item \verb|predict_templates(molecule_id)|: Queries the single-step retrosynthesis model for a given molecule node. Its 50 highest-scoring predictions are shuffled and truncated to at most $k=20$ candidates, each assigned an ID (P1, P2, \ldots) and listing its reactants with per-reactant building block annotations (\verb|building_block: true/false|). Predictions that would create cycles (a reactant is an ancestor of the molecule) or duplicate an existing reaction are automatically filtered out. Within a rollout, subsequent calls for the same molecule reuse the cached candidates with updated filtering at zero additional budget cost.

    \item \verb|expand(molecule_id, prediction_id)|: Applies one or more predictions to the graph, creating new reaction and molecule nodes. Multiple prediction IDs can be comma-separated (e.g., \verb|"P1,P3,P5"|) to create several reaction branches in a single call. Returns the expansion results and the parent node status.

    \item \verb|get_frontier()|: Returns the current set of \texttt{Open} leaf molecules that have not yet been expanded. Also signals when the root molecule has been solved.

    \item \verb|inspect_node(node_id)|: Returns detailed information about any node in the graph, including its SMILES, status (\texttt{Open}/\texttt{Solved}), building block status, parent nodes, and child nodes.
\end{itemize}

\textbf{Chemistry tools.}
These tools provide domain-specific heuristics to help the agent prioritize among candidates. Unlike memory tools, chemistry tools take SMILES strings as input rather than node IDs:
\begin{itemize}[nosep,leftmargin=*]
    \item \verb|get_molecule_info(molecules)|: Returns molecular properties including molecular weight, logP, number of atoms, and ring counts for a list of molecules.

    \item \verb|search_building_blocks(molecules)|: Checks whether given molecules are available in the building block database and returns their availability status.

    \item \verb|calculate_synthetic_accessibility(molecules)|: Computes the synthetic accessibility (SA) score \cite{ertl2009sa} for each molecule, ranging from 1 (easy to synthesize) to 10 (difficult).

    \item \verb|calculate_scscore(molecules)|: Computes the SCScore synthetic complexity \cite{coley2018scscore} for each molecule, ranging from 1 (simple) to 5 (complex).
\end{itemize}

\section{Related Work}
\label{sec:related}

Multi-step retrosynthesis planning methods broadly divide into two streams: \emph{search-based} methods that ground every step in a validated template set, and \emph{model-based} methods that leverage learned representations to generate or predict routes directly.
Our work falls within the search-based paradigm: we use a template-grounded single-step model for chemical validity, but replace the fixed search algorithm with an RL-trained LLM agent that learns its own search strategy.

\textbf{Search-based (template-grounded) methods.}
The dominant paradigm combines a trained single-step retrosynthesis model, typically an MLP classifier over reaction templates \cite{segler2017neural}, with a search algorithm over the resulting AND-OR graph.
Retro* \cite{chen2020retro} introduced neural-guided A* search with a learned value function estimating the cost-to-go for each molecule.
PDVN \cite{liu2023pdvn} improved this with dual value networks that separately predict synthesizability and cost.
EG-MCTS \cite{hong2023egmcts} uses experience from past searches to train a value network for Monte Carlo tree search.
3N-MCTS \cite{segler2018planning} combined MCTS with three neural networks for rollout-based evaluation.
GNN-Retro \cite{han2022gnnretro} improves molecule cost estimation by using a graph neural network that aggregates information from structurally similar molecules.
DESP \cite{yu2024desp} proposed bidirectional search, interleaving top-down retrosynthetic expansion with bottom-up forward synthesis from building blocks, guided by a learned synthetic distance function.
Retro-fallback \cite{tripp2024retrofallback} introduced probabilistic route reliability through stochastic OR nodes.
Beyond node-level heuristics, several works incorporate domain-specific insights into improve multi-step retrosynthesis planning. \citet{mo2021clustering} learn pathway-level representations that rank and cluster retrosynthesis routes by their strategic quality, while \citet{roh2026strategies} reformulate the single-step prediction itself at a higher level of abstraction, separating strategic disconnections from tactical functional group choices to reduce the effective search space.
These methods achieve high solve rates within large iteration budgets (up to 500 single-step model calls) but encode \emph{fixed} search strategies: the heuristic is always a pretrained value function, and the expansion order is determined by the algorithm, not by per-target reasoning.

\textbf{Template-free and generative methods.}
Beyond template-based classifiers \cite{segler2017neural,chen2021localreactivity,coley2017similarity}, template-free methods generate reactants directly.
The Molecular Transformer \cite{schwaller2019molecular} and augmented NLP models \cite{tetko2020state} frame retrosynthesis as sequence-to-sequence translation, while RootAligned \cite{zhong2022rootaligned} enforces a one-to-one mapping between product and reactant tokens by aligning them to a shared root atom.
BatGPT-Chem \cite{yang2025batgptchem} trains an LLM on chemical instruction data for single-step prediction.
LLM-Syn-Planner \cite{wang2025llmsynplanner} uses an LLM to mutate retrieved routes into multi-step plans for new targets.
These approaches can leverage broad chemical knowledge but typically lack the structural guarantees of template-grounded methods.

\textbf{LLM agents for retrosynthesis.}
Recent work has explored LLMs as \emph{agentic planners} that interact with chemistry tools through multi-turn conversations.
Retro-R1 \cite{liu2025retror1} trains a Qwen2.5-7B agent with PPO to select molecules and reactions through a \texttt{CALL/SELECT} interface, achieving state-of-the-art pass@1 success rates.
However, Retro-R1 operates in a \emph{linear, no-backtracking mode}: the agent commits irrevocably to each reaction selection, with no ability to revisit alternative branches.
Its state representation is implicit: the agent must infer the search state from the conversation history.
AOT* \cite{song2025aotstar} combines an LLM with AND-OR tree search, using the LLM to guide node selection within a traditional search framework.
In contrast, our agent has full control over the search process and learns its own strategy end-to-end through RL.

\section{Limitations and Discussion}
\label{app:limitations}

\textbf{Practicality of the generated routes.}
Our evaluation follows the standard template-based protocol: a route is considered valid if every step applies a verified reaction template and all leaves terminate at purchasable building blocks, and quality is measured by success rate, search budget, and route length.
Because all routes pass through the same validated template set and building-block catalog, the routes \ours produces are chemically valid by construction, and the comparison to baselines is on equal footing.
However, this notion of validity does not capture full synthetic practicality: factors such as reaction yield, regio- and stereoselectivity, reaction conditions, and protecting-group strategy are not modeled, and a shorter or template-valid route is not necessarily the most practical one in the laboratory.
These considerations affect all template-based planners equally and depend on information absent from the benchmarks, which provide only target molecules and, for USPTO-190, reference routes.
A consultation with expert chemists found the generated routes generally reasonable, but a systematic, large-scale assessment of practical utility, ideally with experimental validation, remains an open problem for the field and an important direction for future work.

\textbf{Inference cost relative to traditional search.}
\ours relies on LLM inference at every search step, which is substantially more expensive than the lightweight value-network or policy lookups used by traditional search algorithms such as Retro* and PDVN.
A single planning trajectory issues tens of LLM calls and consumes thousands of tokens, whereas a classical planner evaluates its scoring function in milliseconds.
This overhead is the price of the explicit, inspectable reasoning that drives our gains, and we regard it as acceptable in the retrosynthesis setting: planning is a one-time, offline computation of seconds to minutes per target, negligible against the days to weeks required to validate a route experimentally.
Nonetheless, for latency-sensitive or very high-throughput screening, traditional search remains more economical, and reducing LLM planning cost (e.g., through smaller distilled policies or selective invocation) is a worthwhile direction.

\section{Training Details}
\label{app:training}

\textbf{Training hyperparameters.}
\cref{tab:hyperparams} lists the full training hyperparameters for \ours.

\begin{table}
\centering
\caption{Training hyperparameters for \ours with \texttt{Qwen3-4B-Instruct-2507}. The model is trained with full fine-tuning on $8{\times}$H200 GPUs using the slime framework with Megatron and SGLang.}
\label{tab:hyperparams}
\small
\begin{tblr}{
    colspec = {l c},
    row{1} = {font=\bfseries},
}
\toprule
    Parameter & Value \\
    \midrule
    \SetCell[c=2]{l} \emph{Optimization} \\
    Learning rate & $1 \times 10^{-6}$ (constant) \\
    Optimizer & Adam, $\beta_1=0.9$, $\beta_2=0.98$ \\
    Weight decay & 0.01 \\
    Gradient clipping & 1.0 \\
    \midrule[dashed]
    \SetCell[c=2]{l} \emph{RL algorithm} \\
    Advantage estimator & GSPO \\
    Clip ratio & $3.5 \times 10^{-4}$ \\
    KL coefficient & 0.001 \\
    KL loss type & Low-variance KL \\
    Entropy coefficient & 0.0 \\
    \midrule[dashed]
    \SetCell[c=2]{l} \emph{Batching} \\
    Batch size & 32 problems $\times$ 8 rollouts = 256 \\
    Off-policy TIS steps & 2 \\
    TIS clip & 2.0 \\
    Global batch size & 128 \\
    Dynamic oversampling & 48 \\
    \midrule[dashed]
    \SetCell[c=2]{l} \emph{Episode configuration} \\
    Max turns per episode & 90 \\
    Max total tokens & 56,000 \\
    Max response length & 8,192 \\
    Rollout temperature & 1.0 \\
    Max template predictions & 20 \\
    \midrule[dashed]
    \SetCell[c=2]{l} \emph{Reward} \\
    Success / failure reward & $+1.0$ / $-1.0$ \\
    Budget penalty & 0.02 per call above 20, capped at 0.3 \\
    Depth penalty & 0.02 per step above 20, capped at 0.3 \\
    Tool error penalty & $-0.05$ \\
    No-tool-call penalty & $-0.05$ \\
    \midrule[dashed]
    \SetCell[c=2]{l} \emph{Infrastructure} \\
    Tensor parallelism & 2 \\
    SGLang engines & 4 (2 GPUs each) \\
    SGLang context length & 65,536 \\
    SGLang memory fraction & 0.8 \\
    Tool server instances & 16 \\
    GPUs & $8{\times}$H200 \\
\bottomrule
\end{tblr}
\end{table}

\begin{figure}[t]
\centering
\includegraphics[width=\textwidth]{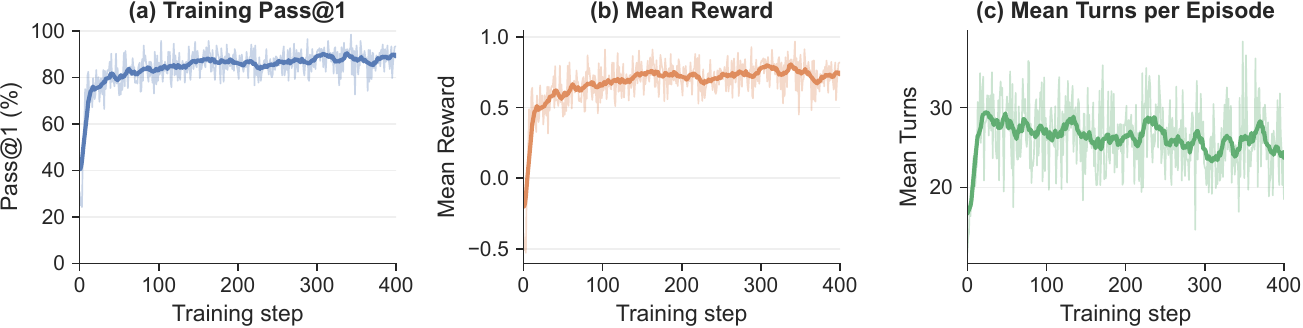}
\caption{Training dynamics of \ours over 400 steps on USPTO-Full targets.
Smoothed trend lines are overlaid on raw per-step values.}
\label{fig:training-curves}
\end{figure}

\textbf{Training curves.}
\label{app:training-curves}
\cref{fig:training-curves} shows the training dynamics of \ours over 400 steps.
Pass@1 on the training set rises rapidly from $\sim$30\% to $\sim$80\% within the first 50 steps and continues to improve gradually, reaching $\sim$90\% by step 400.
Mean reward follows a similar trajectory, stabilizing around 0.75.
Mean turns per episode increases from $\sim$12 to $\sim$27, reflecting the agent's ability to sustain longer, more productive search episodes as it learns.

\section{Evaluation Protocol}
\label{app:eval-protocol}

We evaluate using the standalone eval pipeline with the same tool server as training, with greedy decoding (temperature 0). Each rollout runs for up to 120 turns and a 120K token budget, more generous than training (90 turns, 56K tokens) to allow extra exploration. Following the iterative protocol, each experiment issues successive rollouts for a target until a route is found or the search budget is exhausted; we run 10 independent experiments per target and report the mean and standard deviation, with stochasticity coming entirely from shuffled prediction and frontier orderings in the tool server, not from LLM sampling.
A rollout terminates when the root molecule is solved, when maximum turns are reached, when no successful expansion occurs for several consecutive turns (no progress), when the frontier is exhausted (dead frontier), when the token budget is exhausted, or when a single turn exceeds the response limit of 8,192 tokens due to repetition.
Reaching the maximum route depth of 30 reaction steps blocks only the offending branch and does not by itself end the rollout.

\section{Additional Experiments}
\label{app:additional}

\subsection{Same-Backbone Comparison with Retro-R1}
\label{app:same-backbone}

To isolate the contribution of our structured-memory interface from the choice of base model, we reproduce the Retro-R1 two-action (\texttt{CALL/SELECT}) interface on the same \texttt{Qwen3-4B-Instruct-2507} backbone used by \ours, holding the model and RL training procedure fixed and varying only the interface.
We use the released Retro-R1 code, the step-200 checkpoint, and the same iterative evaluation protocol on USPTO-190, reporting five independent runs.
As shown in \cref{tab:same-backbone}, the Retro-R1 interface on the \texttt{Qwen3-4B-Instruct-2507} backbone reaches $45.05\pm0.29$\% pass@1, $8.21$ points below \ours (53.26\%) on the same backbone and even below the original Retro-R1 result of 50.00\% on \texttt{Qwen2.5-7B-Instruct-1M}.
Thus moving the Retro-R1 interface onto our backbone does not by itself improve performance; the gain comes from the interface and learned policy rather than the base model.
Together with the zero-shot result of 4.42\% pass@1 (\cref{tab:ablation}), this shows the interface does not help without RL training, and that the gains are not attributable to the choice of base model.

\begin{table}
\centering
\caption{Same-backbone comparison on USPTO-190 (five-run mean$\pm$std). Retro-R1's two-action interface is reproduced on the same \texttt{Qwen3-4B-Instruct-2507} backbone used by \ours, holding the model and RL training fixed.}
\label{tab:same-backbone}
\small
\begin{tblr}{
    colspec = {l l c ccc},
    row{1} = {bg=gray!25, font=\bfseries},
    row{3} = {bg=gray!10},
}
\toprule
    Method & Base model & Pass@1 & $N=50$ & $N=100$ & $N=200$ \\
    \midrule
    Retro-R1 & \texttt{Qwen2.5-7B-Instruct-1M} & 50.00{\scriptsize$\pm$0.00} & 64.63{\scriptsize$\pm$1.68} & 71.32{\scriptsize$\pm$1.23} & 78.11{\scriptsize$\pm$1.68} \\
    Retro-R1 & \texttt{Qwen3-4B-Instruct-2507} & 45.05{\scriptsize$\pm$0.29} & 50.63{\scriptsize$\pm$3.08} & 60.32{\scriptsize$\pm$1.96} & 71.37{\scriptsize$\pm$3.10} \\
    \midrule[dashed]
    \ours & \texttt{Qwen3-4B-Instruct-2507} & \textbf{53.26}{\scriptsize$\pm$0.69} & \textbf{58.42}{\scriptsize$\pm$1.34} & \textbf{70.53}{\scriptsize$\pm$1.93} & \textbf{78.63}{\scriptsize$\pm$1.27} \\
\bottomrule
\end{tblr}
\end{table}

\subsection{Reward Sensitivity}
\label{app:reward-sweep}

We study the sensitivity of \ours to its reward shaping from two perspectives: the relative weighting of the budget penalty $\delta_N$ and depth penalty $\delta_D$, and the overall magnitude of both penalties.
All configurations are otherwise identical to \ours and trained from the same base model.
\cref{tab:reward} reports USPTO-190 success rates as mean$\pm$std over five runs.

On the relative weighting, keeping both penalty terms is best (53.26\%), and dropping either one lowers pass@1 only modestly (budget-only 51.16\%, depth-only 50.84\%), so the two terms are complementary but performance is robust to their balance.
On the magnitude, doubling both penalty rates and caps lowers pass@1 to 48.32\% but the budget curve converges ($N=200$ rises to 75.58\%, approaching the full model's 78.63\%).
Harsher penalties make the agent more conservative, trading first-attempt success for budget efficiency without removing the underlying ability to solve targets.

\begin{table}[b]
\centering
\caption{Reward sensitivity on USPTO-190 (five-run mean$\pm$std). Relative weighting moves pass@1 by only a few points; doubling the penalty magnitude trades first-attempt success for a converging budget curve.}
\label{tab:reward}
\small
\begin{tblr}{
    colspec = {l c ccc},
    row{1} = {bg=gray!25, font=\bfseries},
    row{3,5} = {bg=gray!10},
}
\toprule
    Reward configuration & Pass@1 & $N=50$ & $N=100$ & $N=200$ \\
    \midrule
    Both penalties (\ours) & \textbf{53.26}{\scriptsize$\pm$0.69} & \textbf{58.42}{\scriptsize$\pm$1.34} & \textbf{70.53}{\scriptsize$\pm$1.93} & \textbf{78.63}{\scriptsize$\pm$1.27} \\
    Budget-only (w/o depth penalty) & 51.16{\scriptsize$\pm$2.59} & 54.95{\scriptsize$\pm$1.42} & 64.84{\scriptsize$\pm$2.39} & 73.79{\scriptsize$\pm$1.01} \\
    Depth-only (w/o budget penalty) & 50.84{\scriptsize$\pm$2.65} & 53.68{\scriptsize$\pm$1.97} & 64.63{\scriptsize$\pm$2.72} & 74.11{\scriptsize$\pm$0.69} \\
    Both, doubled magnitude & 48.32{\scriptsize$\pm$2.09} & 56.00{\scriptsize$\pm$2.37} & 68.32{\scriptsize$\pm$0.94} & 75.58{\scriptsize$\pm$0.80} \\
\bottomrule
\end{tblr}
\end{table}

\section{Out-of-Distribution Analysis}
\label{app:ood}

We quantify the distribution shift between the ChEMBL-1000 evaluation set and the USPTO-Full training targets in terms of three properties (\cref{tab:ood}):
\begin{itemize}[nosep,leftmargin=*]
    \item \textbf{Scaffold novelty:} we extract the Bemis-Murcko scaffold of each molecule and mark it novel if it does not appear among the training targets; 96.2\% of ChEMBL-1000 scaffolds are novel, versus 74.8\% for USPTO-190.
    \item \textbf{Similarity to training:} the mean maximum Tanimoto similarity (Morgan fingerprints, radius 2, 2048 bits) of each target to the training set is 0.345 for ChEMBL-1000 versus 0.483 for USPTO-190, and 85\% of ChEMBL-1000 targets fall below the USPTO-190 median.
    \item \textbf{Molecular complexity:} synthetic accessibility is comparable between the two sets (median 3.18 for ChEMBL-1000 versus 3.55 for USPTO-190).
\end{itemize}
Together, these show the shift is driven by scaffold novelty and dissimilarity to training rather than by raw complexity.
Despite the strong scaffold-level shift, \ours solves the 962 scaffold-novel ChEMBL-1000 targets at 73.8\% pass@1 and 84.2\% at $N\le500$ (810/962), essentially matching the full-set rates (74.4\% pass@1, 84.6\% at $N\le500$).
This indicates the agent does not rely on memorized scaffold-to-template associations but instead learns a policy driven by distribution-independent signals such as building-block proximity.

\begin{table}
\centering
\caption{Distribution shift of ChEMBL-1000 relative to USPTO-190 and the training targets.}
\label{tab:ood}
\small
\begin{tblr}{
    colspec = {l c c},
    row{1} = {bg=gray!25, font=\bfseries},
    row{3} = {bg=gray!10},
}
\toprule
    Property & USPTO-190 & ChEMBL-1000 \\
    \midrule
    Scaffold-novel vs.\ training (\%) & 74.8 & 96.2 \\
    Mean max.\ Tanimoto to training & 0.483 & 0.345 \\
    SA score (median) & 3.55 & 3.18 \\
\bottomrule
\end{tblr}
\end{table}

\section{Failure Case Analysis}
\label{app:failures}

We categorize every failed target on USPTO-190 and ChEMBL-1000 by the per-episode termination reason logged by the environment (\cref{tab:failures}).
The environment records four mutually exclusive termination reasons:
\begin{itemize}[nosep,leftmargin=*]
    \item \texttt{max\_turns}: the episode reaches the turn limit (120) without solving the root molecule;
    \item \texttt{no\_progress}: the agent makes no successful expansion for five consecutive turns and is stopped early;
    \item \texttt{dead\_frontier}: every open molecule has been expanded yet none yields a solved route;
    \item \texttt{token\_limit}: the episode exhausts the 120K-token budget.
\end{itemize}
Reaching the maximum route depth of 30 reaction steps only blocks the offending branch rather than ending the episode, so such attempts are subsumed within \texttt{max\_turns} or \texttt{no\_progress}.
On both benchmarks, budget exhaustion dominates: \texttt{max\_turns} and \texttt{no\_progress} together account for $\sim$96\% of failures on USPTO-190 and $\sim$94\% on ChEMBL-1000, while \texttt{dead\_frontier} and \texttt{token\_limit} together remain under 7\%.
Failures are therefore concentrated on hard, deep targets that exhaust the search budget rather than on the agent running out of candidates or hitting action-space limits.
We also note that failed targets receive a similar number of first-step predictions as solved ones (15.0 vs.\ 17.9 on USPTO-190; 16.6 vs.\ 17.0 on ChEMBL-1000); failures thus arise from exhausting the search budget on hard, deep targets, not from a lack of initial candidates.

\begin{table}[b]
\centering
\caption{Termination reason among failed targets (\% of failures; $n=27$ on USPTO-190, $n=154$ on ChEMBL-1000).}
\label{tab:failures}
\small
\begin{tblr}{
    colspec = {l c c},
    row{1} = {bg=gray!25, font=\bfseries},
    row{3,5} = {bg=gray!10},
}
\toprule
    Termination reason & USPTO-190 & ChEMBL-1000 \\
    \midrule
    \texttt{max\_turns} & 74.1 & 57.8 \\
    \texttt{no\_progress} & 22.2 & 35.7 \\
    \texttt{dead\_frontier} & 3.7 & 3.2 \\
    \texttt{token\_limit} & 0.0 & 3.2 \\
\bottomrule
\end{tblr}
\end{table}

\section{Example Transcripts}
\label{app:transcripts}

We present two solved and two failed episodes from USPTO-190 evaluation.
Each transcript shows the full tool call sequence, where T$n$ denotes the $n$-th turn of the agent-environment interaction.
For each \verb|predict_templates| call, we list the number of returned predictions and the Building Block (BB) count per prediction.
For \verb|expand| calls, we show which predictions the agent selected; $w=k$ denotes the expansion width, i.e., the number of predictions applied in a single call.

\textbf{Solved example 1.}
Solved in 11 turns with 4 predict calls, reaching depth 4 with 18 molecule nodes.

\begin{tcolorbox}[breakable, colback=green!5, colframe=green!60!black, fontupper=\small\ttfamily, title={\small Solved: 11 turns{,} 4 predictions{,} depth 4}]
Target: CC(C)c1ccc(-n2nc(O)c3c(=O)c4ccc(Cl)cc4[nH]c3c2=O)cc1\\[5pt]
T1: predict\_templates(M1) -> 20 predictions\\
\phantom{T1: }P1: 2 reactants (1 BB), P2--P20: 1 reactant each (0 BB)\\[4pt]
T2: expand(M1, [P1,P3,P5,P7,P11]) w=5\\[4pt]
T3: get\_frontier() -> [M2, M3, M4, M5, M6]\\[4pt]
T4: predict\_templates(M2) -> 20 predictions\\
\phantom{T4: }P6: 2 reactants (1 BB), P10: 2 reactants (1 BB), P12: 2 reactants (1 BB)\\
\phantom{T4: }P1--P5, P7--P9, P11, P13--P20: 1 reactant each (0 BB)\\[4pt]
T5: expand(M2, [P6,P10,P12]) w=3 \textrm{// selects all 3 with BB}\\[4pt]
T6: get\_frontier() -> [M9, M10, M11, M12]\\[4pt]
T7: predict\_templates(M9) -> 20 predictions\\
\phantom{T1: }P1--P20: 1 reactant each (0 BB)\\[4pt]
T8: expand(M9, [P17,P20]) w=2\\[4pt]
T9: get\_frontier() -> [M13, M14]\\[4pt]
T10: predict\_templates(M13) -> 20 predictions\\
\phantom{T10: }P1: 2 reactants (1 BB), P2: 2 reactants (1 BB), P14: 2 reactants (2 BB)\\
\phantom{T10: }remaining: 1 reactant each (0 BB)\\[4pt]
T11: expand(M13, [P1,P14]) w=2 -> root SOLVED
\end{tcolorbox}

The agent selects predictions with building block reactants: at T5, it picks exactly the 3 predictions with BB annotations out of 20 candidates. At T11, it picks P14 which has 2 BB reactants, completing the route.

\textbf{Solved example 2.}
Solved in 14 turns with 5 predict calls, reaching depth 5 with 22 molecule nodes.

\begin{tcolorbox}[breakable, colback=green!5, colframe=green!60!black, fontupper=\small\ttfamily, title={\small Solved: 14 turns{,} 5 predictions{,} depth 5}]
Target: CCOC(=O)c1nc(N2CC[C@H](NC(=O)c3nc(C(F)(F)F)c(CC)[nH]3)[C@H](OC)C2)sc1C\\[5pt]
T1: predict\_templates(M1) -> 16 predictions\\
\phantom{T1: }P1(1BB), P2(1BB), P3(1BB), P5(1BB), P10(1BB), P11(1BB), P12(1BB): 2 reactants\\
\phantom{T1: }P4, P6--P9, P13--P16: 1 reactant (0 BB)\\[4pt]
T2: expand(M1, [P1,P3,P5,P11,P12]) w=5 \textrm{// all with BB}\\[4pt]
T3: get\_frontier() -> [M2, M4, M6, M8, M9, M10]\\[4pt]
T4: predict\_templates(M9) -> 20 predictions\\
\phantom{T4: }P17: 2 reactants (1 BB); rest: 1 reactant (0 BB)\\[4pt]
T5: expand(M9, [P17]) w=1 \textrm{// only one with BB}\\[4pt]
T6: get\_frontier() -> [M11]\\[4pt]
T7: predict\_templates(M11) -> 16 predictions\\
\phantom{T7: }P1: 3 reactants (2 BB), P2(1BB), P6(1BB), P7(1BB), P8(1BB), P10(1BB), P14(1BB)\\[4pt]
T8: expand(M11, [P1,P3,P10,P14]) w=4\\[4pt]
T9: get\_frontier() -> [M19]\\[4pt]
T10: predict\_templates(M19) -> 11 predictions\\
\phantom{T10: }P7: 2 reactants (1 BB); rest: 1 reactant (0 BB)\\[4pt]
T11: expand(M19, [P7]) w=1 \textrm{// only one with BB}\\[4pt]
T12: get\_frontier() -> [M20]\\[4pt]
T13: predict\_templates(M20) -> 8 predictions\\
\phantom{T13: }P1: 1 reactant (1 BB), P3: 2 reactants (2 BB)\\[4pt]
T14: expand(M20, [P1,P3]) w=2 -> root SOLVED
\end{tcolorbox}

The agent adapts its expansion width based on prediction quality: wide ($w=5$) at the root where many predictions have BB reactants, narrow ($w=1$) when only one prediction has BBs, and moderate otherwise.

\textbf{Failed example 1.}
Failed after 112 turns, 38 predict calls, depth 30, 95 molecule nodes.

\begin{tcolorbox}[breakable, colback=red!5, colframe=red!40, fontupper=\small\ttfamily, title={\small Failed: 112 turns{,} 38 predictions{,} depth 30 (no progress)}]
Target: C[C@H](c1ccccc1)N1C[C@]2(C(=O)OC(C)(C)C)C=CC[C@@H]2C1=S\\[5pt]
T1: predict\_templates(M1) -> 8 predictions [P1(1BB), P2(1BB), P3(1BB), ...]\\[4pt]
T2: expand(M1, [P1,P6,P8]) w=3\\[4pt]
T3: get\_frontier() -> [M2, M3, M4, M5]\\[4pt]
T4: predict\_templates(M2) -> 13 predictions [P1(2BB), ...]\\[4pt]
T5: expand(M2, [P1,P4,P5,P13]) w=4\\[4pt]
T6: get\_frontier() -> [M11, M12, M13, ...]\\[4pt]
T7: predict\_templates(M11) -> 19 predictions [0 BB each]\\[4pt]
T8: expand(M11, [P7,P14]) w=2\\[4pt]
... (92 more turns: predict -> expand -> get\_frontier cycle)\\[4pt]
Termination: no\_progress (5 consecutive turns without successful expand)
\end{tcolorbox}

\textbf{Failed example 2.}
Failed after 103 turns, 35 predict calls, depth 30, 88 molecule nodes.

\begin{tcolorbox}[breakable, colback=red!5, colframe=red!40, fontupper=\small\ttfamily, title={\small Failed: 103 turns{,} 35 predictions{,} depth 30 (no progress)}]
Target: CCCC[C@@H](C(=O)N1CCC[C@H]1C(=O)O)[C@@H](F)C(=O)OC\\[5pt]
T1: predict\_templates(M1) -> 13 predictions [0 BB]\\[4pt]
T2: expand(M1, [P5,P6,P13]) w=3\\[4pt]
T3: get\_frontier() -> [M2, M3, M4, ...]\\[4pt]
T4: predict\_templates(M2) -> 17 predictions [P1(1BB), P2(1BB)]\\[4pt]
T5: expand(M2, [P1,P6,P12]) w=3\\[4pt]
... (83 more turns)\\[4pt]
Termination: no\_progress
\end{tcolorbox}

Both failed episodes follow the dominant failure pattern quantified in \cref{app:failures}: the agent sustains a long predict-expand-frontier cycle but cannot reach building blocks on these hard, deep targets, eventually triggering early \texttt{no\_progress} termination.

\end{document}